\pdfoutput=1

\documentclass[11pt]{article}

\usepackage[final]{acl}

\usepackage{times}
\usepackage{latexsym}

\usepackage[T1]{fontenc}
\usepackage{enumitem}

\usepackage[utf8]{inputenc}

\usepackage{microtype}

\usepackage{inconsolata}

\usepackage{graphicx}
\usepackage{tabularx}
\usepackage{pifont}   
\usepackage[table]{xcolor} 
\usepackage{booktabs}
\usepackage{colortbl,caption,subcaption}
\usepackage{stfloats} 
\usepackage{array}

\usepackage{etoolbox} 

\newif\ifinappendix
\inappendixfalse

\pretocmd{\appendix}{\inappendixtrue}{}{}

\makeatletter
\let\origaddcontentsline\addcontentsline
\renewcommand{\addcontentsline}[3]{%
  \origaddcontentsline{#1}{#2}{#3}%
  \ifinappendix
    \ifstrequal{#1}{toc}{\origaddcontentsline{apx}{#2}{#3}}{}%
  \fi
}
\makeatother

\newcolumntype{L}[1]{>{\raggedright\arraybackslash}p{#1}}
\definecolor{headercolor}{RGB}{217,225,242}   
\definecolor{rowcolorA}{RGB}{248,248,248}     
\definecolor{dgreen}{RGB}{0,180,0}   
\definecolor{dred}{RGB}{200,0,0}       

\newcommand{\modelname}{MADIAVE}
\newcommand{\llamaname}{Llama-3.2-11B-Vision-Instruct}
\newcommand{\Qwenname}{Qwen-2.5-VL-7B}
\newcommand{\phiname}{Phi-3.5-Vision-Instruct}

\usepackage{amsmath}

\renewcommand{\arraystretch}{1.3}    
\newcolumntype{C}{>{\centering\arraybackslash}X}

\title{MADIAVE: Multi-Agent Debate for Implicit Attribute Value Extraction}

\author{
  Wei-Chieh Huang$^1$ \quad Cornelia Caragea$^1$ \\
  $^1$University of Illinois Chicago \\
  \texttt{\{whuang80, cornelia\}@uic.edu}
}

\usepackage{chngcntr}   

\begin{document}
\maketitle

\begin{abstract}
Implicit Attribute Value Extraction (AVE) is essential for accurately representing products in e-commerce, as it infers latent attributes from multimodal data. Despite advances in multimodal large language models (MLLMs), implicit AVE remains challenging due to the complexity of multidimensional data and gaps in vision-text understanding. In this work, we introduce \textsc{\modelname}, a multi-agent debate framework that employs multiple MLLM agents to iteratively refine inferences. Through a series of debate rounds, agents verify and update each other's responses, thereby improving inference performance and robustness. Experiments on the ImplicitAVE dataset demonstrate that even a few rounds of debate significantly boost accuracy, especially for attributes with initially low performance. We systematically evaluate various debate configurations, including identical or different MLLM agents, and analyze how debate rounds affect convergence dynamics. Our findings highlight the potential of multi-agent debate strategies to address the limitations of single-agent approaches and offer a scalable solution for implicit AVE in multimodal e-commerce.
\end{abstract}

\section{Introduction}
\label{sec:introduction}

Technological advancements have radically reshaped consumer behavior, establishing e-commerce as an indispensable part of modern life. Recent studies \citep{mohdhar2021future} indicate that e-commerce will continue to drive future application development and thrive in an increasingly digital landscape, underscoring its ever-growing importance. However, accurately identifying product attributes remains a significant challenge \citep{zhang2024stronger}, as incorrect or imprecise labels can lead to customer dissatisfaction and a loss of trust and loyalty \citep{Lin2021Pam}. In response, Attribute Value Extraction (AVE) has emerged as a promising approach to enhance product representation and categorization \citep{Zheng:2018, Yan:2021, yang-etal-2023-mixpave}. Owing to the complexity and enormous volume of data, AVE has attracted significant research interest \citep{hu2025hypergraph, gong2025visual}, leading to the development of various methods to more effectively address this challenge.

\begin{figure}[t]
  \includegraphics[width=\columnwidth]{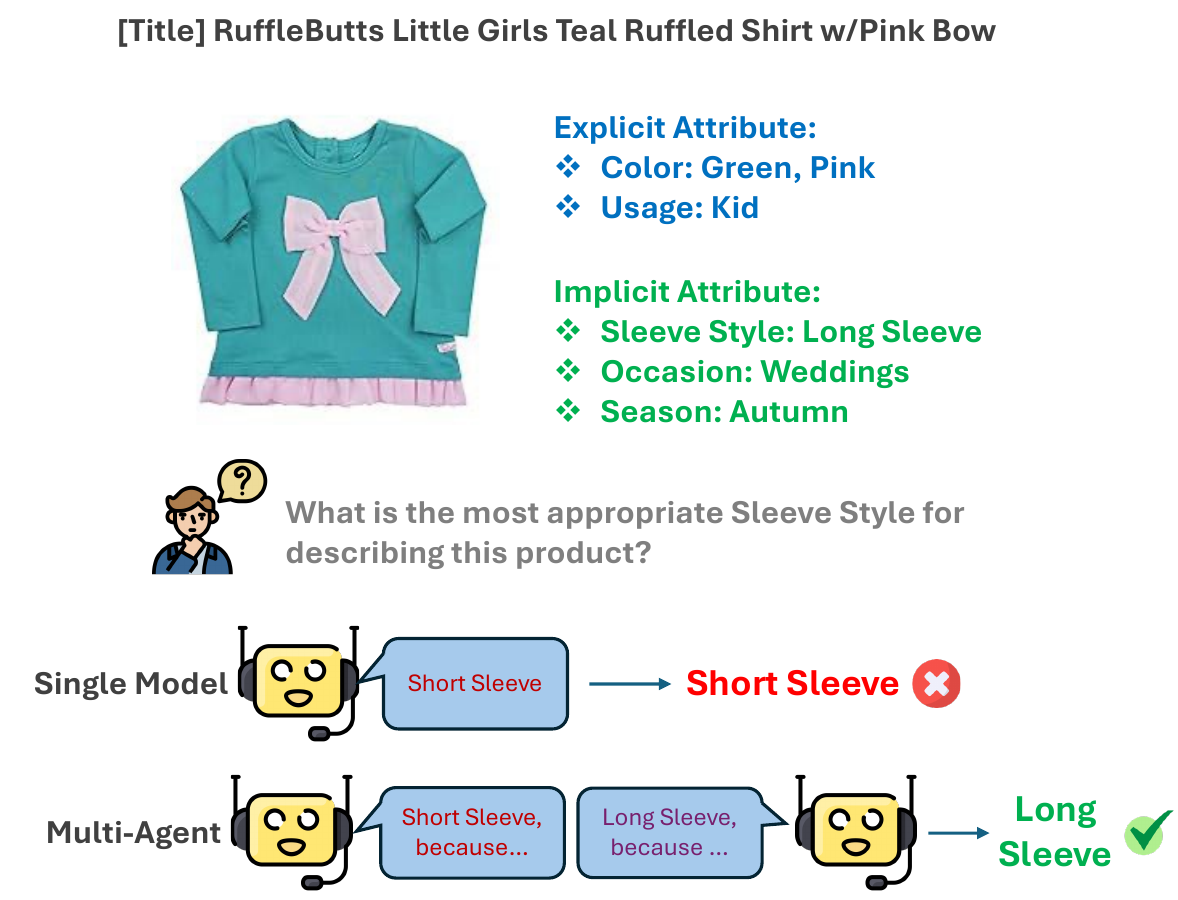}
    \caption{Illustration of the differences between explicit and implicit attributes. A single model may sometimes produce incorrect inferences, whereas a multi-agent setting can potentially provide more accurate reasoning and lead to correct inferences.}
  \label{fig:fig0}
  \vspace{-5mm}
\end{figure}

AVE can generally be classified into two categories: \textit{\textbf{(1) Explicit}} \citep{sabeh2024exploring, su2025taclr} and \textit{\textbf{(2) Implicit}} \citep{gong2024multi, jia2024mujo}. Explicit approaches extract attributes that are directly mentioned in the textual description, whereas implicit approaches infer attributes from visual cues and the textual information, as illustrated in Figure~\ref{fig:fig0}. Due to the inherent challenges of the task, most research has focused on explicit AVE. For instance, some recent studies \citep{10.1145/3523227.3547395, loughnane-etal-2024-explicit} have proposed model architectures based on transformers and named-entity recognition models to address explicit AVE tasks. However, a key limitation of these approaches is that they rely solely on textual similarity for label extraction rather than inferring attributes from underlying content or images, potentially overlooking latent attributes. Consequently, researchers have begun exploring implicit AVE methods. Some studies \citep{zhang-etal-2023-pay, Luo:2023, zou2024eivenefficientimplicitattribute} propose advanced methods and architectures aimed at reducing the need for expensive labeled data in implicit AVE. Yet, these methods can still be costly to train, often require labeled data, and achieve only moderate inference performance. Addressing these challenges remains a key area for further research.

To address these challenges, we propose a novel framework, \textsc{\modelname}, which integrates textual and visual reasoning across multiple agents to enhance overall performance in a fully {\em zero-shot setting} with multimodal large language models (MLLMs). Inspired by the success of large language models (LLMs) \citep{zoullm, DBLP:journals/corr/abs-2502-16804} in general inference tasks \citep{xu-etal-2024-identifying, he-etal-2024-llm}, and the robust capabilities of agents \citep{DBLP:journals/corr/abs-2506-18959, DBLP:journals/corr/abs-2507-09477, DBLP:journals/corr/abs-2506-09420}, we posit that multi-round, multi-agent debate and reasoning \citep{li2025towards, zhang2025web} can refine model outputs, guiding them toward consensus and ultimately achieving correct inferences for the implicit AVE task. To better understand agent behavior during debate, we systematically analyze performance and convergence in debates among both identical and different agents. We expect this framework not only to advance the state of the art in implicit AVE, but also to serve as a representative approach for other multimodal tasks.

Our contributions are summarized as follows:

\begin{itemize}
    \item \textbf{Novel Framework for Implicit AVE:} We present a novel framework that integrates reasoning capabilities of multi-agents through debate to enhance inference performance. To best of our knowledge, this is the first study to investigate multi-agent debate mechanism into multimodal implicit AVE task. 
    \item \textbf{Comprehensive Evaluation of Debate Configurations:} We systematically evaluate various debate configurations, including those involving identical, diverse, open-source, and closed-source MLLMs in a zero-shot setting. This analysis provides valuable insights into how different MLLMs contribute to and impact each other's performance.
    \item \textbf{In-depth Analysis of Debate Convergence:} We present a detailed examination of debate statistics and reasoning consensus to understand how agents adopt, refine, or reject each other’s viewpoints during the debate, and how final convergence is ultimately achieved.
\end{itemize}

\section{Related Works}
\label{sec:related works}

\subsection{Explicit and Implicit AVE}
\label{sec:ImplicitAVE}

\looseness=-1
Early stages of AVE research primarily focused on explicit extraction methods. For example, early approaches \citep{Zheng:2018, Wang:2020} utilized conventional long short-term memory (LSTM) networks and Bidirectional Encoder Representations from Transformers (BERT) to tackle the AVE task. However, the inherent complexity of multimodal data limited the accuracy of these methods. Although recent research has introduced advanced techniques \citep{yang-etal-2023-mixpave, sabeh2024empiricalcomparisongenerativeapproaches}, enhanced encoders \citep{wang-etal-2022-smartave}, and innovative architectures \citep{brinkmann2024extractgptexploringpotentiallarge, loughnane-etal-2024-explicit} for explicit AVE, these approaches often overlook potential attributes derivable from visual cues and textual information. Alternatively, the dataset and framework focus on the task of mixed evidence for both explicit and implicit AVE in e-commerce products, making the evaluation of implicit performance unreliable \citep{wang2022smartave, liu2023multimodal, wang2023mpkgac}.

As a result, research has shifted toward implicit AVE. Current studies have employed multimodal generative frameworks \citep{zhang-etal-2023-pay} and efficient multimodal LLMs \cite{zou2024eivenefficientimplicitattribute} to improve inference accuracy in implicit AVE tasks. Despite these advances, implicit AVE remains in its early stages, generally achieving only moderate accuracy and revealing significant potential for improvement. To address this gap, we propose the \textsc{\modelname} framework, which enhances inference performance by leveraging multi-agent and multi-round debate and reasoning \citep{wu2025multi}. Our approach integrates insights from multiple agents to better capture latent attributes in multimodal data, thereby advancing the state of the art in implicit AVE.

\begin{figure*}[t]
  \centering
  \includegraphics[width=\textwidth]{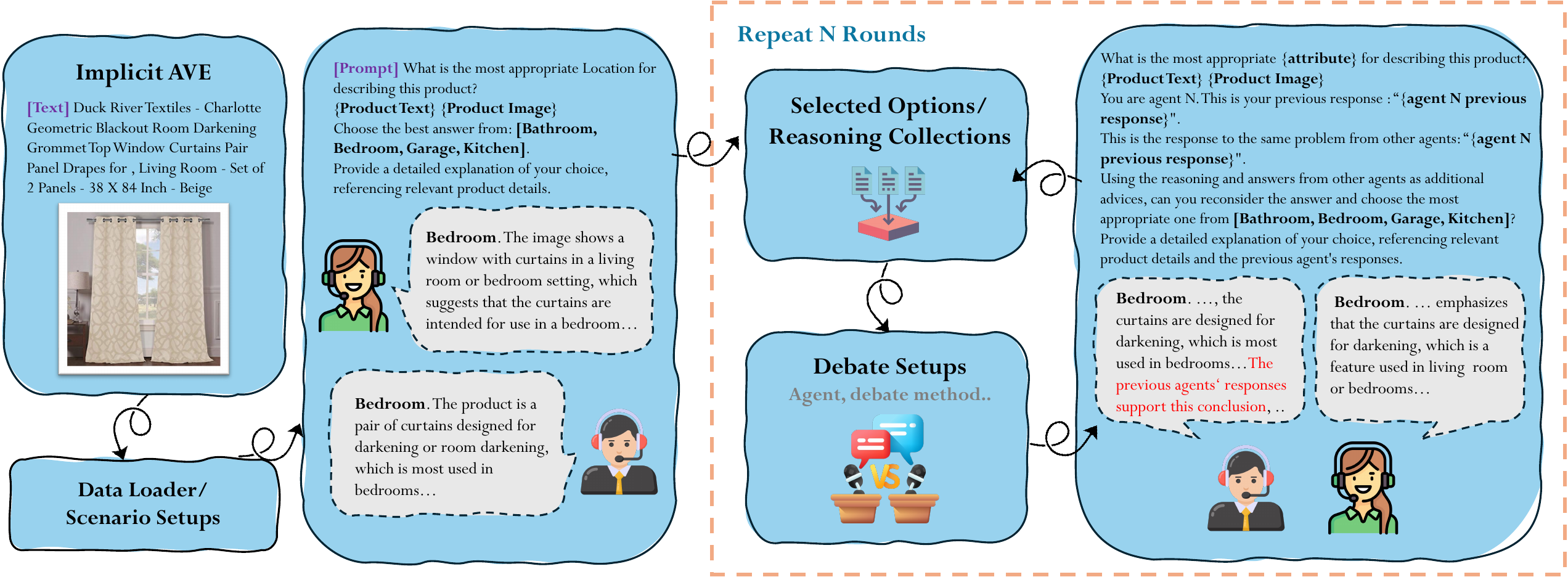}
  \caption{\textbf{Overview of the \textsc{\modelname} framework}. The ImplicitAVE is ingested by a data loader that applies scenario steps to initialize agent roles and trigger multi-round communication and debate. After each round, the message transcript is recorded and appended to the context, serving as additional input for the next round.}
  \label{fig:fig1}
\end{figure*}

\subsection{MLLM, LLM and Debate}
\label{sec:MLLMDebate}

Recent research in LLMs and MLLMs has demonstrated impressive inference capabilities in tasks such as visual question answering \citep{hu2024bliva} and cross-modal retrieval \citep{10843094}. Despite significant progress in processing multimodal data, achieving high inference accuracy remains challenging due to the complexity of multidimensional data. For example, although previous work on implicit AVE has utilized LLMs and MLLMs for inference,with models like EIVEN \citep{zou2024eivenefficientimplicitattribute} reducing reliance on labeled data, they can still produce inaccurate results under certain conditions. Moreover, while LLMs have shown strong performance with minimal in-context examples \citep{blume-etal-2023-generative}, these approaches are often limited by expensive training processes and only moderate overall performance.

As a consequence, relying on a single LLM may not yield optimal results and can introduce bias. Researchers have explored methods to integrate the intelligence of multiple agents on tasks such as presidential debates \citep{liu2024llmpotusscoreframeworkanalyzing}, debate simulations \citep{Taubenfeld_2024}, stance and argument analysis \citep{heinisch-etal-2024-tell}, and approaches to improve factuality and reasoning \citep{du2023improving}. Recent investigations reveal that a multi-agent debate strategy can further enhance performance by enabling models to converge on more accurate answers in open-ended scenarios \citep{chan2023chatevalbetterllmbasedevaluators, khan2024debatingpersuasivellmsleads}. However, these studies have focused exclusively on textual input, neglecting the visual component and the integration of visual cues with textual information during inference. In addition, most research has examined debates among identical LLMs under limited input conditions, leaving the potential for collaboration between different MLLMs largely unexplored.

To address these gaps, we propose the \textsc{\modelname} framework as a novel solution for the implicit AVE task. Our approach leverages multi-round, multi-agent debate among MLLMs to integrate both textual and visual reasoning, thereby improving inference accuracy for implicit AVE tasks.

\begin{table*}[ht]
  \centering
  \scriptsize
  \rowcolors{2}{rowcolorA}{white}
  \caption{\label{tab:prompt-config}
    Prompt configurations of the initial and debate rounds for MLLMs .}
  \vspace{-0.4em}
  \begin{tabularx}{\textwidth}{L{0.28\textwidth} X}

    \rowcolor{headercolor}
      \textbf{Condition Setting} 
    & \textbf{Prompt Text} \\
    \midrule
    Initial — 1\textsuperscript{st} Round
    & \emph{What is the most appropriate \{attribute\} for describing this product?}\\
    & \emph{Context: Category: \{category\}, Title: \{title\}. Choose the best answer from: \{options\}.}\\
    \addlinespace[4pt]
    Debate — 2\textsuperscript{nd} to 5\textsuperscript{th} Round
    & \emph{What is the most appropriate \{attribute\} for describing this product?}\\
    & \emph{Context: Category: \{category\}, Title: \{title\}.}\\
    & \emph{You are agent \{agent\_num\}. Your previous response: “\{previous\_response\_1\}”.}\\
    & \emph{Response from other agents: “\{previous\_response\_2\}”.}\\
    & \emph{Using other agents’ reasoning as advice, reconsider and choose the best option from: \{options\}.}\\
    \addlinespace[2pt]
    \bottomrule
  \end{tabularx}
  \vspace{-2mm}
  
\end{table*}

\section{MADIAVE Framework}
\label{sec:Implicit AVE through Multiagent Debate}

We present our \textsc{\modelname} framework for the implicit AVE task in Section~\ref{sec:Multiagent Debate}. This section details the overall inference and debate process, explaining how multiple agents interact to refine their inferences. In Section~\ref{sec:Consensus between Agents in Debate}, we examine the behavior and convergence of the debate over multiple rounds and how the agents collectively reach consensus. The detailed \textsc{\modelname} framework, experimental setup, and workflow are illustrated in Figure~\ref{fig:fig1}. The prompts for the initial round and subsequent debate rounds are provided in Table~\ref{tab:prompt-config}.

\subsection{Multi-agent Debate}
\label{sec:Multiagent Debate}

In our context, an “agent” refers to an MLLM that functions much like a human, with the capacity to update its answers dynamically. For instance, when addressing an implicit AVE task such as determining a product’s shape from its image and text, an MLLM might infer possible answers like “heart,” “oval,” or “rectangle.” Although this attribute is not explicitly stated in the text, it is derived through both visual and textual cues. While a single MLLM might generate a tentative inference, the reliability of the answer is enhanced when multiple agents interact and converge on the same conclusion. Examining each agent’s reasoning and explanations provides insight into diverse perspectives and reveals how inferences are continuously refined. Moreover, our \textsc{\modelname} framework adopts a zero-shot setting to circumvent extensive training requirements and reliance on a training set, allowing us to evaluate whether MLLMs have sufficient general understanding and reasoning abilities regarding the product. In this multi-agent debate procedure, each MLLM emulates human-like reasoning and revises its answer based on the other agent's insights, ultimately leading to a more robust and reliable solution for the implicit AVE task.

To take advantage of this reasoning process and improve the inference capabilities of MLLMs, the \textsc{\modelname} framework first sets up two to five agents to independently analyze the implicit AVE dataset \cite{zou2024implicitaveopensourcedatasetmultimodal} by addressing questions such as ``What is the length of this product?''. Each agent selects an answer from a predefined option list and provides a justification for its choice, indicating whether its decision is based on visual evidence, textual cues, or both. This corresponds to the first inference block in Figure~\ref{fig:fig1}. 

\looseness=-1
After the initial round of inference, the agents proceed to subsequent rounds where they receive not only the original product image and text but also the responses from both agents, including each answer and its accompanying reasoning. For instance, in rounds 2 through 5, agent 1 will receive the image, text, its own previous response, and agent 2’s previous response. This iterative process enables each agent to incorporate the reasoning, consensus, or disagreements from the previous round as additional context, ultimately leading to a more reliable and refined answer. This process is shown in the selected option and reasoning collection and the debate block in Figure~\ref{fig:fig1}.

\subsection{Consensus between Agents in Debate}
\label{sec:Consensus between Agents in Debate}

During debates among agents, several outcomes are possible. When an agent’s answer is challenged, its confidence may decrease as it is exposed to alternative perspectives, prompting it to reconsider its evaluation of the product. This interaction may lead an agent to modify its answer, reach consensus with other agents, or even attempt to persuade other agent that its original response is correct. Over multiple rounds of debate, we expect the agents to gradually converge toward a shared consensus; however, achieving consensus does not necessarily guarantee that the final answer is correct for the implicit AVE task. We show empirical findings of the consensus during the debate in Appendix~\ref{sec:appendix1}

\begin{table*}[ht]
  \centering
  \caption{\label{table2}Overview of scenario configurations. This table details the agent setups for each scenario.}
  \vspace{-6pt}
  \scriptsize
  \begin{tabularx}{\textwidth}{@{\extracolsep{\fill}}
      >{\centering\arraybackslash}p{8mm}
      >{\centering\arraybackslash}p{40mm}  
      >{\centering\arraybackslash}p{40mm}  
      X                                    
      @{}}
    \hline
    \textbf{Scenario} & \textbf{Agent1} & \textbf{Agent2} & \textbf{Scenario Detail} \\
    \hline
    1 & GPT-4o            & GPT-4o            & Same LLM (Closed-Source Model, Zero-Shot) Debate \\
    2 & \llamaname         & \llamaname         & Same LLM (Open-Source Model, Zero-Shot) Debate \\
    3 & \phiname          & \phiname           & Same LLM (Open-Source Model, Zero-Shot) Debate \\
    4 & \llamaname        & GPT-4o            & Different LLM (Closed, Open-Source, Zero-Shot) Debate \\
    5 & \llamaname     & \phiname          & Different LLM (Open-Source Model, Zero-Shot) Debate \\
    6 & \Qwenname       & \Qwenname        & Same LLM (Open-Source Model, Zero-Shot) Debate \\
    7 & GPT-o1            & GPT-o1            & Same LLM (Closed-Source Model, Zero-Shot) Debate \\
    8 & Claude-3.5-Sonnet & Claude-3.5-Sonnet & Same LLM (Closed-Source Model, Zero-Shot) Debate \\
    \hline
  \end{tabularx}
  \vspace{-3mm}
\end{table*}

\section{Experiment}
\label{sec:experiment}

In this section, we describe the experimental setup. Details about the benchmark are provided in Section~\ref{sec:ImplicitAVE Dataset for Debate}. The debate scripts are provided in Section~\ref{app:debate-scripts}. The debate constructions and scenarios are described in Section~\ref{sec:MLLM agents for Debate} and Section~\ref{sec:Debate Scenarios}.

\begin{table*}[!b] 
  \centering
  \scriptsize
  \setlength{\tabcolsep}{6pt}
  \renewcommand{\arraystretch}{1.08}
  \caption{Domain level data statistics for the evaluation set of the ImplicitAVE dataset.}
  \label{table1}
  \begin{tabularx}{\textwidth}{@{}lcccX@{}}
    \hline
    \textbf{Domain} & \textbf{\# Data} & \textbf{\# Attribute} & \textbf{\# Value} & \textbf{Attributes} \\
    \hline
    Clothing    & 226 & 4 & 23 & Neckline, Sleeve Style, Length, Shoulder Style \\
    Footwear    & 317 & 5 & 29 & Athletic Shoe Style, Boot Style, Shaft Height, Heel Height, Toe Style \\
    Jewelry\&GA & 220 & 3 & 20 & Pattern, Material, Shape \\
    Home        & 457 & 8 & 45 & Material, Special Occasion, Location, Animal Theme, Season, Shape, Size, Attachment Method \\
    Food        & 390 & 5 & 41 & Flavor, Form, Candy Variety, Occasion, Container \\
    \hline
  \end{tabularx}
\end{table*}

\subsection{ImplicitAVE Benchmark}
\label{sec:ImplicitAVE Dataset for Debate}
The quality of an implicit AVE dataset is essential for achieving accurate inference and reasoning. However, we have identified several issues with current implicit AVE datasets:
\textit{\textbf{(1) implicit AVE research is still in its early stages, so few datasets are available}};
\textit{\textbf{(2) many datasets are not publicly available}};
\textit{\textbf{(3) some datasets conflate explicit and implicit AVE}} \citep{yang2024eave, wang2022smartave}. To minimize the impact of noise, we rely on a high-quality benchmark ImplicitAVE dataset \citep{zou2024implicitaveopensourcedatasetmultimodal}, derived from the MAVE dataset \citep{MAVE:2022}. Its evaluation set encompasses five domains—Clothing, Footwear, Jewelry\&GA, Home, and Food—for a total of 1,610 samples. Each domain includes five to eight attributes and spans 23 to 45 values, providing high-quality images and accompanying text. This evenly distributed evaluation set is ideally suited for the multi-agent debate process and serves as an excellent testbed for the \textsc{\modelname} framework. Further details, including the number of attributes and attribute values, can be found in Table~\ref{table1}.

\subsection{MLLM agents for Debate}
\label{sec:MLLM agents for Debate}
    

Our \textsc{\modelname} framework adopts a \emph{zero-shot} setting throughout the paper. This design choice is deliberate. While few-shot prompting and parameter-efficient fine-tuning methods (e.g., LoRA) can often improve task accuracy, they typically require task- and domain-specific examples that may not reflect the full diversity of real-world e-commerce profiles, queries, and product catalogs. In practice, collecting and curating high-quality labeled demonstrations at scale is expensive, and training on only thousands of examples, compared to the billions of interactions and heterogeneous distributions encountered in real deployments, can introduce undesirable bias and lead to overfitting to a narrow slice of user behaviors. This risk is particularly serious in e-commerce, where long-tail preferences, rapidly changing inventories, and shifting seasonal trends can cause distribution shift and reduce robustness under realistic conditions. To prioritize broad generalizability and minimize additional supervision, we therefore evaluate debate agents purely via zero-shot inference.

Concretely, we instantiate debate agents with six state-of-the-art MLLMs that span both closed-source and open-source models:
\texttt{GPT-4o},
\texttt{\llamaname},
\texttt{\phiname},
\texttt{\Qwenname},
\texttt{Claude-3.5-Sonnet}, and
\texttt{GPT-o1}.
This selection allows us to assess whether the debate dynamics induced by \textsc{\modelname} are consistent across model families and capabilities, rather than being an artifact of a single backbone or a tuned prompt.

\paragraph{Few-shot prompting and instruction-tuning.}
We emphasize that few-shot prompting and instruction-tuned backbones are \emph{promising} and can further boost absolute performance. However, such gains are conditioned on the availability and representativeness of curated examples: demonstrations that work well for a subset of products, user personas \citep{wu2025psg}, or browsing patterns may not transfer to unseen catalogs or novel preference signals. Because the central goal of this work is to evaluate \textsc{\modelname} as a \emph{general} multi-agent debate framework, one that can be applied to new domains and catalogs with minimal additional supervision, we focus on the zero-shot settings as the most conservative and widely applicable evaluation setting.

\subsection{Debate Scenarios and Experiments}
\label{sec:Debate Scenarios}
We thoroughly investigate a variety of debate scenarios related to multimodal data and AVE tasks. By varying the debate agents, we capture a broad range of possibilities. Specifically, we define eight distinct scenarios, as detailed in Table~\ref{table2}. To quantify the benefit of debate, we include a control experiment of majority vote and an ablation that measures performance sensitivity to the number of agents and the number of debate rounds.

\section{Debate Results}
\label{sec:Debate Results}

\subsection{Debate Result and Analysis}

\subsubsection{Domain-Level Result}

Table~\ref{tab:baseline comparison} reports domain-level and overall accuracy for three settings: prior baselines, single-model inference, and the \textsc{\modelname} self-debate framework. Relative to their single-model counterparts, all listed models improve under \textsc{\modelname}. Among single model runs, \texttt{GPT-o1} has the highest overall score at 86.27, closely followed by \texttt{GPT-4o}. In terms of by domain, \texttt{GPT-4o} leads \emph{Clothing} and \emph{Jewelry\&GA}, while \texttt{GPT-o1} leads \emph{Footwear} and \emph{Food} and achieves the best \emph{Overall} accuracy. The only exception is \emph{Home}, where the earlier \texttt{GPT-4V} baseline remains the top score at 89.93. Overall, the table shows that self-debate reliably boosts performance across models, with \texttt{GPT-4o} and \texttt{GPT-o1} providing the strongest domain and aggregate results under \textsc{\modelname}. To test the stability of the performance, we run the experiment three times and the result can be found in the Appendix~\ref{app:ave-stability}.

\begin{figure}[t]
  \includegraphics[width=\columnwidth]{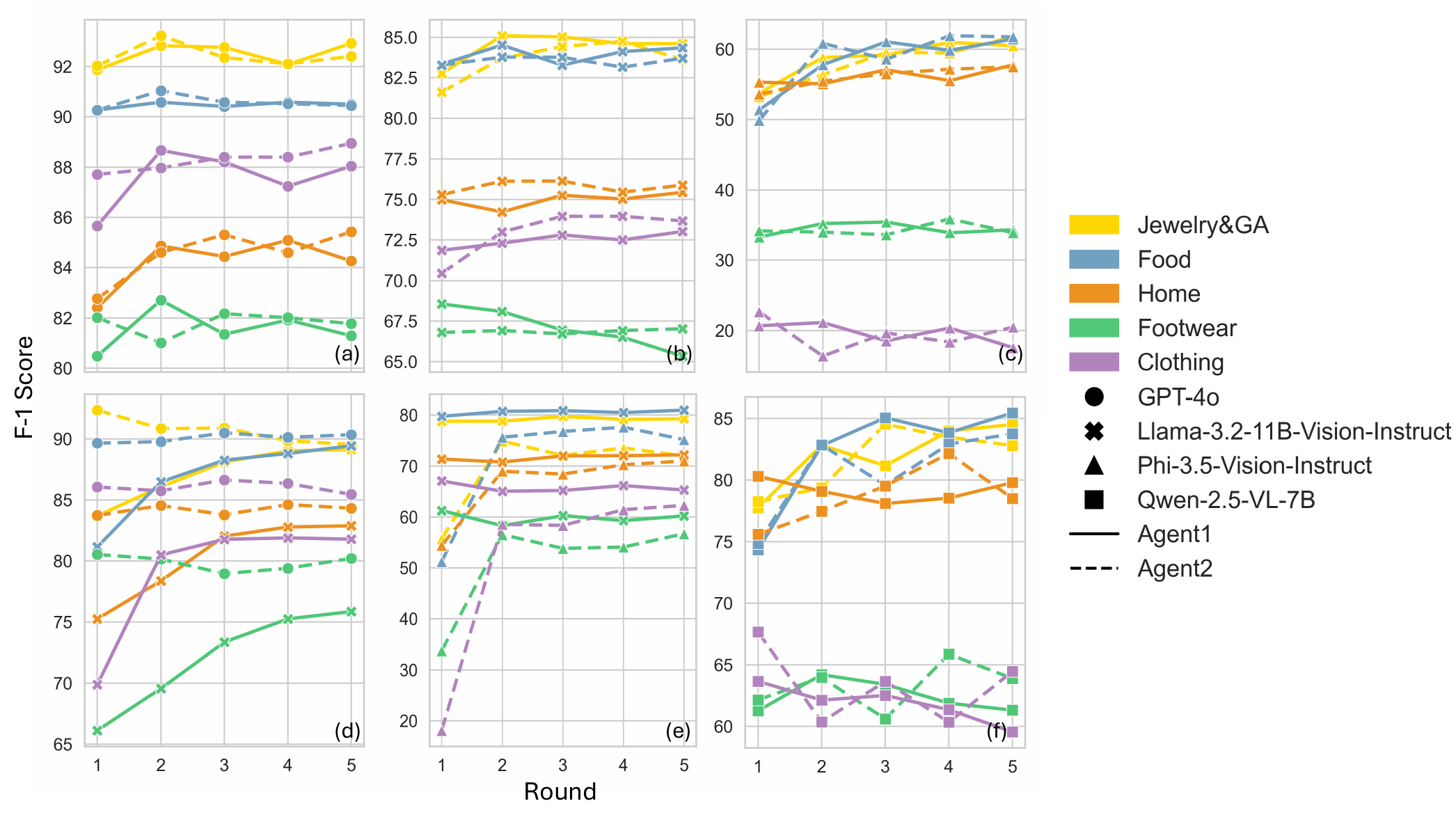}
  \caption{Domain-level F1 trajectories during debate, reported per agent across rounds. (a) Scenario 1: GPT-4o debating with GPT-4o; (b) Scenario 2: \llamaname~debating with \llamaname; (c) Scenario 3: \phiname~debating with \phiname; (d) Scenario 4: \llamaname~debating with GPT-4o; (e) Scenario 5: \llamaname~debating with \phiname (f) Scenario 6: \Qwenname~debating with \Qwenname.}
  \label{fig:Domain-Level Result}
\end{figure}

We present domain-level debate results in Figure~\ref{fig:Domain-Level Result}, highlighting performance differences among the MLLMs in various scenarios. In the same MLLM debate in Figures~\ref{fig:Domain-Level Result}(a), \ref{fig:Domain-Level Result}(b), \ref{fig:Domain-Level Result}(c), and \ref{fig:Domain-Level Result}(f), we observe that the F1 score increases during the first one or two rounds. In particular, for \texttt{GPT-4o}, the debate results show a general improvement in F1 score from 0.2\% to 3\% across all domains. However, after round two, the accuracy decreases or fluctuates within a limited range, indicating that although the debate converges, additional rounds do not yield further improvements. However, the inference performance after two to five rounds remains superior to that of a single agent.

Similarly, we observe an initial increase in inference performance after one round of debate for \texttt{\llamaname}, with improvements ranging from 0.12\% to 2.55\%. However, this gain is accompanied by some noise and uncertainty, as evidenced by a 0.75\% decrease in the Home domain. Overall, \texttt{\llamaname}'s improvement is more moderate and stable compared to \texttt{GPT-4o}, and as the debate continues, its performance fluctuates within a limited range, mirroring \texttt{GPT-4o}'s behavior. On the other hand, for \texttt{\phiname} and \texttt{\Qwenname}, F1 improves after the first debate round, yet further rounds bring no additional gains and instead cause performance to fluctuate within a narrow and fixed range.

\begin{table*}[ht]
  \centering
  \scriptsize
  \begin{tabularx}{370pt}{l l c c c c c c c}
    \hline
    \textbf{Method} & \textbf{Language Model} & 
    \textbf{Clothing} & 
    \textbf{Footwear} & 
    \textbf{Jewelry\&GA} & 
    \textbf{Food} & 
    \textbf{Home} &
    \textbf{Overall} \\
    \hline
    \multicolumn{8}{l}{\textbf{Previous Baseline}\ \citep{zou2024implicitaveopensourcedatasetmultimodal}} \\
    BLIP-2 & FlanT5XXL-11B & 55.31 & 55.21 & 82.72 & 71.02 & 71.33 & 67.39 \\
    InstructBLIP & FlanT5XXL-11B & 62.83 & 63.41 & 83.18 & 73.58 & 73.96 & 71.49 \\
    LLaVA-1.5 & Vicuna-13B & 49.12 & 63.72 & 81.81 & 76.15 & 80.31 & 71.86 \\
    Qwen-VL & Qwen-7B & 68.14 & 62.14 & 84.09 & 76.92 & 73.96 & 70.86 \\
    GPT & GPT-4V & 77.43 & 81.39 & 90.45 & 90.77 & \textbf{\textcolor{dgreen}{89.93}} & 86.77 \\
    \multicolumn{8}{l}{\textbf{Single Model}} \\
    GPT & GPT-4o & 85.66 & 80.48 & 91.87 & 90.26 & 82.41 & 85.68 \\
    Llama-3.2 & \llamaname & 70.44 & 66.8 & 81.62 & 83.28 & 74.98 & 75.64  \\
    Phi-3.5 & \phiname & 20.67 & 33.27 & 53.24 & 49.86 & 53.6 & 44.02 \\
    Qwen2.5 & \Qwenname & 64.16 & 61.83& 77.72 & 74.87 & 75.71 & 71.43 \\
    Claude & Claude-3.5-Sonnet & 71.68 & 69.40 & 79.55 & 82.05 & 76.57 & 76.21 \\ 
    GPT & GPT-o1 & 84.96 & 81.08 & 91.36 & 90.26 & 83.81 & 86.27 \\ 
    \multicolumn{8}{l}{\textbf{\textsc{\modelname} Framework (Multi-Agent Debate)}} \\
    GPT & GPT-4o & \textbf{\textcolor{dgreen}{88.95}} & 82.71 & \textbf{\textcolor{dgreen}{93.23}} & 91.04 & 85.78 & 87.91 \\
    Llama-3.2 & \llamaname & 73.96 & 68.09 & 85.1 & 84.52 & 76.14 & 77.50 \\
    Phi-3.5 & \phiname & 22.67 & 35.87 & 60.99 & 61.93 & 57.79 & 49.98 \\
    Qwen2.5 & \Qwenname & 61.95 & 65.93 & 85.00 & 85.13 & 81.84 & 77.14 \\
    Claude & Claude-3.5-Sonnet & 76.10 & 66.88 & 80.45 & 84.61 & 78.77 & 77.70 \\ 
    GPT & GPT-o1 & 88.50 & \textbf{\textcolor{dgreen}{83.28}} & 92.27 & \textbf{\textcolor{dgreen}{91.80}} & 86.88 & \textbf{\textcolor{dgreen}{88.32}} \\ 
    \hline
  \end{tabularx}
  \caption{\label{tab:baseline comparison}
    The \textsc{\modelname} framework is compared to previous baseline methods and single model inference on the ImplicitAVE dataset. The aggregated system performance is evaluated across five individual domains as well as overall accuracy. The \textbf{\textcolor{dgreen}{bold green}} entries indicate the overall best performance among the methods.}
  \vspace{-1mm}
\end{table*}

These findings consistently indicate that the \textsc{\modelname} framework improves inference performance relative to single model approaches, with one or two rounds of debate yielding notable improvement, whereas additional rounds do not provide further benefits. In many instances, agents converge on incorrect inferences, possibly because MLLMs tend to prioritize an opposing agent’s advice over their own initial judgment, thus creating confusion. Moreover, the results show that \texttt{GPT-4o} outperforms the other models with F1-scores ranging from 80\% to 93\%, peaking at 93.23\% in the Jewelry \& GA domain at round two. \texttt{\llamaname} achieves moderate performance, ranging from 65\% to 85\%, whereas \texttt{\phiname} exhibits the lowest performance. \texttt{GPT-4o} demonstrates a more stable and pronounced improvement compared to \texttt{\llamaname} and \texttt{\Qwenname}, mirroring real-life scenarios in which knowledgeable participants lead the discussion toward better consensus while less-informed individuals may inadvertently lead to suboptimal convergence or wrong directions.

Additionally, a domain-specific analysis reveals that the Jewelry \& GA domain generally exhibits superior inference performance compared to other domains. In contrast, the Clothing and Footwear domains consistently demonstrate lower performance. This observation suggests that either the MLLM agents have a training bias favoring certain domains, or that the Clothing and Footwear domain is inherently more challenging for all MLLM agents—potentially even for human reviewers. We analyze attribute-level inference performance and present the result in Appendix~\ref{sec:attribute-level result}.

\subsection{Convergence Statistic}

\begin{figure}[t]
  \includegraphics[width=\columnwidth]{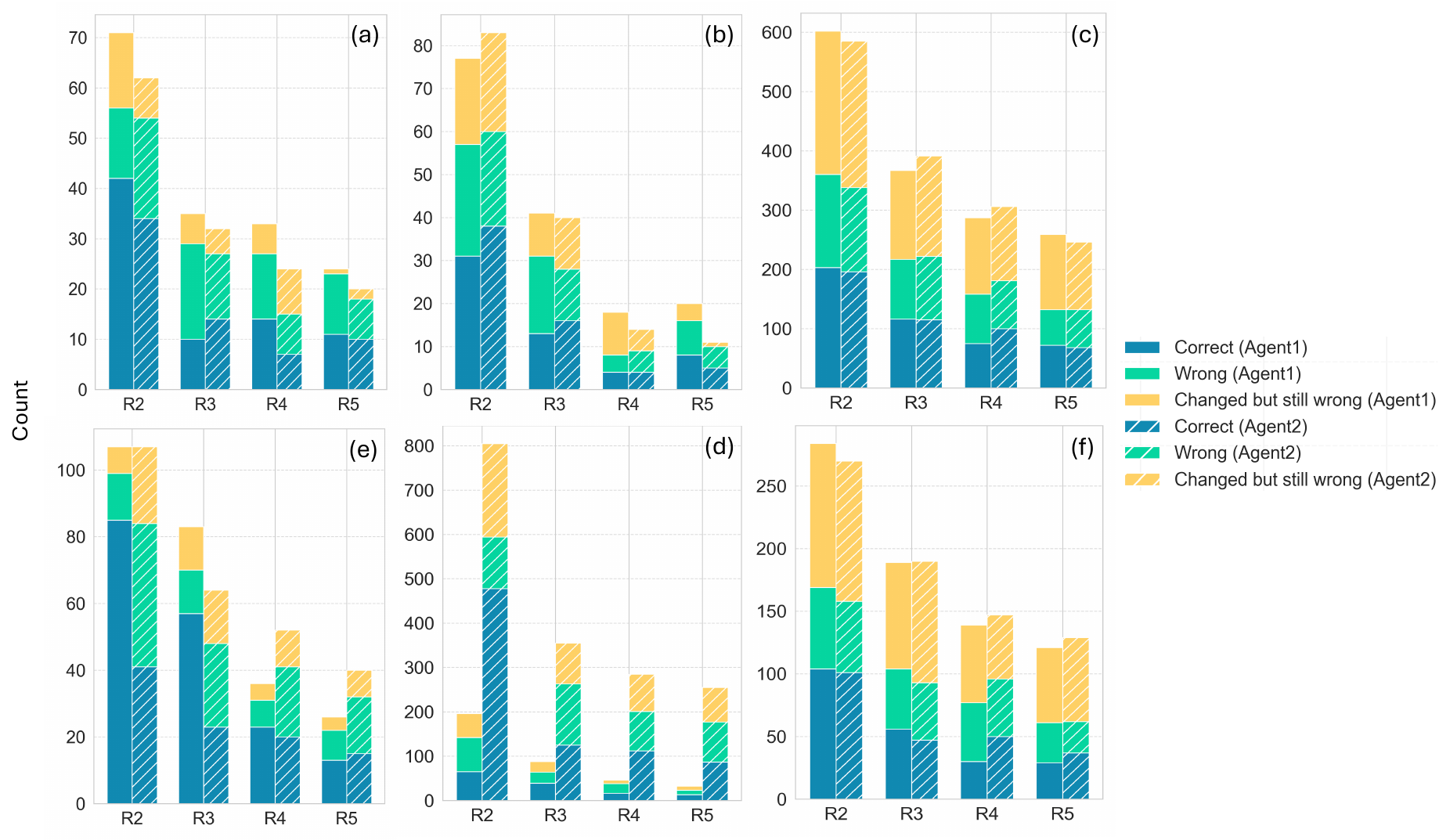}
  \caption{Debate statistics for (a) Scenario 1: GPT-4o debating with GPT-4o; (b) Scenario 2: \llamaname~debating with \llamaname; (c) Scenario 3: \phiname~debating with \phiname; (d) Scenario 4: \llamaname~debating with GPT-4o; (e) Scenario 5: \llamaname~debating with \phiname; (f) Scenario 6: \Qwenname~debating with \Qwenname.}
  \label{fig:Convergence Stats}

\end{figure}

Detailed convergence information is provided in Figures~\ref{fig:Convergence Stats}. In the same MLLM debates involving \texttt{GPT-4o}, \texttt{\llamaname}, \texttt{\phiname}, and the Qwen faimily \texttt{\Qwenname}, as shown in Figures~\ref{fig:Convergence Stats}(a), \ref{fig:Convergence Stats}(b), \ref{fig:Convergence Stats}(c), \ref{fig:Convergence Stats}(f), respectively, we observe that the agents exhibit the most changes during round two, the first round of debate, with the frequency of changes decreasing in subsequent rounds. For example, in Figure~\ref{fig:Convergence Stats}(a) for \texttt{GPT-4o}, the total number of changes drops from 70 to 20, indicating that the debate steadily converges until both agents reach consensus on the inference. Additionally, the number of cases where an agent changes its answer yet still arrives at an incorrect inference decreases over time, suggesting that agents increasingly tend to follow other agent’s selection rather than its own options. However, although there is an initial improvement in correct inferences during round two, the number of worsened inferences begins to exceed the number of improved ones by round three, implying that the convergence process can sometimes deviate in the wrong direction. In such cases, agents appear to become confused by flawed reasoning, failing to properly assess whether to adopt correct or incorrect reasoning. A similar pattern is observed in Figures~\ref{fig:Convergence Stats}(b) and \ref{fig:Convergence Stats}(c). It is also noteworthy that weaker agents, due to their limited ability to evaluate reasoning, tend to change their answers more frequently. For instance, when \texttt{\phiname} debates with itself, both agents change their answers in 600 out of a total of 1,610 instances.This number remains around 300 even after four rounds of debate.

\begin{table*}[t]
\centering
\small
\setlength{\tabcolsep}{5pt}
\renewcommand{\arraystretch}{1.12}
\caption{The performance comparison on GPT-4o: single inference, majority vote (10 independent runs), and two-agent debate (2 agents, 5 rounds).}
\label{tab:mv}
\resizebox{120mm}{!}{%
\begin{tabular}{l c c c c c c}
\toprule
\textbf{Method (GPT-4o)} & \textbf{Clothing} & \textbf{Footwear} & \textbf{Jewelry \& GA} & \textbf{Food} & \textbf{Home} & \textbf{Overall} \\
\midrule
Single Inference & 85.66 & 80.48 & 91.87 & 90.26 & 82.41 & 85.68 \\
Majority Vote    & 87.17 & \textbf{82.88} & 92.72 & 90.26 & 83.15 & 86.69 \\
Debate (2 agents, 5 rounds) & \textbf{88.95} & 82.71 & \textbf{93.23} & \textbf{91.04} & \textbf{85.78} & \textbf{87.91} \\
\bottomrule
\end{tabular}%
}
\end{table*}

For different MLLM debates, we observe that the convergence patterns differ from those seen in the same agent debates. In Figures~\ref{fig:Convergence Stats}(d) and \ref{fig:Convergence Stats}(e), the stronger agent, acting as a "teacher", often ends up with worsened answers, likely due to the influence of inaccurate reasoning from the weaker "student" agent. In contrast, the student agent consistently corrects more inferences in each round, indicating that it benefits significantly from the teacher's input during the debate. In Figure~\ref{fig:Convergence Stats}(e), although the weaker learner, \texttt{\phiname}, does not exhibit strong reasoning or a clear understanding of the product image and text, it changes its answers approximately 800 times in round 2 and corrects about 500 data points, by the guidance and support of \texttt{\llamaname} model. Conversely, the teacher agent shows more deteriorated answers across rounds, suggesting it benefits less from the exchange and can be pulled off course by the student’s confident but weakly grounded arguments. This indicates a reverse-influence effect: prompts that encourage incorporating peer reasoning may cause the teacher to over-weight unreliable feedback and drift from correct choices. Overall, while the teacher’s guidance is invaluable for improving the weaker agent, the teacher agent may pay an additional performance cost by partially adopting or accepting the student agent’s inaccurate or imprecise reasoning, especially in cases or environment that are not well-defined.

\subsection{The Performance Comparison Between Single Inference, Majority Vote, and Dsebate}
\label{sec:majority vote}
To differentiate the advantages of debate from additional sampling, we aligned the inference budget by executing a \emph{single} \texttt{GPT-4o} agent ten times (equivalent to two agents across five rounds, \(2\times5=10\)) and compared \emph{Single Inference}, \emph{Majority Vote}, and \emph{Debate}.  Table~\ref{tab:mv} demonstrates that debate achieves the highest \emph{Overall} accuracy at 87.91\%, surpassing majority vote at 86.69\% and single inference at 85.68\%.  Domain-specific improvements are uniform, with the most significant increase observed in \emph{Home} (85.78\% compared to 83.15\% majority and 82.41\% single).  Debate enables agents to examine heuristic assumptions and integrate cross-modal evidence (e.g., correlating packing dimensions with visual indicators), while majority vote aggregation of independent executions lacks a framework and procedure for reasoning exchange, resulting in the continuation of initial errors.

\begin{table*}[!t]
\centering
\small
\setlength{\tabcolsep}{5pt}        
\renewcommand{\arraystretch}{1.12} 
\caption{Agents\,$\times$\,Rounds ablation under a fixed compute budget of 10 model calls (interaction depth vs.\ agent breadth)}
\label{tab:ablation}
\resizebox{110mm}{!}{
\begin{tabular}{cccccccc}
\toprule
\textbf{Rounds} & \textbf{Agents} & \textbf{Clothing} & \textbf{Footwear} & \textbf{Jewelry \& GA} & \textbf{Food} & \textbf{Home} & \textbf{Overall} \\
\midrule
5 & 2 & 88.95 & 81.39 & 92.27 & 90.26 & 85.34 & 86.83 \\
4 & 2 & 87.61 & 81.70 & 91.82 & 90.26 & 84.92 & 86.90 \\
3 & 2 & 88.05 & 82.02 & 92.27 & 90.52 & \textbf{85.78} & 87.39 \\
2 & 2 & 88.50 & \textbf{82.71} & \textbf{93.23} & \textbf{91.04} & 84.25 & \textbf{87.42} \\
\midrule
2 & 3 & \textbf{89.82} & 81.39 & 90.91 & 89.74 & 84.46 & 86.77 \\
2 & 4 & 87.17 & 80.44 & 90.91 & 87.69 & 82.71 & 85.22 \\
2 & 5 & 86.28 & 82.02 & 91.82 & 88.98 & 83.81 & 86.15 \\
\bottomrule
\end{tabular}%
}
\end{table*}

\begin{table*}[t]
\centering
\small
\setlength{\tabcolsep}{6pt}
\renewcommand{\arraystretch}{1.12}
\caption{One round debates (R2) and single round (R1) baseline across six agents. Accuracy gains come with higher latency but modest token growth. Efficiency (Gain/\(\Delta t\)) highlights improvement per added second.}
\label{tab:debate-lat-acc}
\resizebox{\textwidth}{!}{%
\begin{tabular}{l c c c c c c c c c c}
\toprule
\multicolumn{1}{c}{\textbf{Model}} &
\multicolumn{1}{c}{\textbf{R1 Lat. (s)}} &
\multicolumn{1}{c}{\textbf{R1 Acc. (\%)}} &
\multicolumn{1}{c}{\textbf{R2 Lat. (s)}} &
\multicolumn{1}{c}{\textbf{R2 Acc. (\%)}} &
\multicolumn{1}{c}{\(\boldsymbol{\Delta}\)\textbf{Acc. (\%)}} &
\multicolumn{1}{c}{\textbf{Lat. \(\times\)}} &
\multicolumn{1}{c}{\textbf{\(\Delta\mathrm{Acc}/\Delta t\) (\% s\(^{-1}\))}} &
\multicolumn{1}{c}{\textbf{Init Tok}} &
\multicolumn{1}{c}{\textbf{R2 Tok}} &
\multicolumn{1}{c}{\textbf{Tok \(\times\)}} \\
\midrule
GPT-4o & 3.72 & 85.68 & 7.95 & 87.42 & +1.74 & 2.14 & 0.41 & 80.31 & 99.60 & 1.24 \\
\llamaname & 4.32 & 75.64 & 8.96 & 77.02 & +1.38 & 2.07 & 0.30 & 79.32 & 97.22 & 1.23 \\
\phiname & 3.29 & 44.02 & 6.84 & 48.63 & +4.61 & 2.08 & 1.30 & 85.17 & 95.33 & 1.12 \\
\Qwenname & 2.32 & 71.43 & 6.08 & 75.90 & +4.47 & 2.62 & 1.19 & 76.55 & 94.32 & 1.23 \\
Claude-3.5-sonnet & 7.16 & 76.21 & 12.77 & 77.20 & +0.99 & 1.78 & 0.18 & 133.25 & 155.96 & 1.17 \\
GPT-o1 & 8.73 & 86.27 & 19.21 & 87.52 & +1.25 & 2.20 & 0.12 & 157.41 & 189.25 & 1.20 \\
\bottomrule
\end{tabular}%
}
\end{table*}

\subsection{Ablations on Agents and Rounds}
\label{sec:ablations}

To study how \textsc{\modelname} trades-off \emph{interaction depth} (rounds) versus \emph{agent breadth} (number of agents), we follow a standard inference-time budget control used in multi-agent LLM systems: we hold the \emph{number of model calls} fixed and vary only how those calls are orchestrated. Concretely, we fix the budget to 10 total calls and compare settings such as two agents over five rounds versus five agents over two rounds. This call-level control isolates the effect of debate structure, whereas token-level matching is difficult to enforce in practice and can be misleading due to variable context growth and heterogeneous reasoning behavior across models. Table~\ref{tab:ablation} shows that with two agents, \textbf{2--3 rounds} yield the strongest performance (Overall 87.42--87.39\%), while extending to 4--5 rounds provides no additional benefit (86.90--86.83\%). Under the same 10-call budget with two rounds, increasing breadth from 2 to 3--5 agents \emph{reduces} accuracy (86.77\%, 85.22\%, 86.15\%). Overall, these results suggest that a small team with brief exchanges best leverages debate: early rounds help reconcile cross-modal evidence, but additional rounds or larger committees introduce redundancy and noise, leading to diminishing returns and occasional drift.

Combined with the call-matched comparison in Section~\ref{sec:majority vote}, these findings indicate that \textsc{\modelname}'s gains persist when controlling for the number of communications. \textsc{\modelname} outperforms both single inference and majority vote under the same call budget, particularly on harder implicit cues (e.g., \emph{Home}). We note that debate typically incurs additional context tokens as transcripts accumulate; this is an inherent trade-off of richer orchestration rather than a confound we can reliably equalize. Practitioners should therefore view \textsc{\modelname} as \emph{call-comparable} to majority vote and decide whether the extra context cost is justified under their latency and token constraints. As a practical default, we recommend \textbf{two agents with two rounds} (optionally three), and scaling up only for difficult slices after cost--latency considerations.



\subsection{Latency Analysis and Accuracy-Cost Trade-offs of Debate}
\label{sec:results-debate-latency}
In a comparison of six agents, the most informative metric is efficiency, defined as gain per additional second \(\Delta\mathrm{Acc}/\Delta t\). Two rounds of debate reveal that weak models has a higher gain, with \texttt{\phiname} at 1.30\%\,s\(^{-1}\) and \texttt{\Qwenname} at 1.19\%\,s\(^{-1}\); more robust models exhibit just moderate returns, such as \texttt{GPT-4o} at 0.41\%\,s\(^{-1}\) and \texttt{GPT-o1} at 0.12\%\,s\(^{-1}\). Absolute accuracy improvements range from \(+0.99\) to \(+4.61\)\%, although latency increases by a factor of \(1.78\) to \(2.62\) compared to a single round. Therefore, the overall quality–time trade-off favors more on weak models under time constraints. Improvements at the category level focus on the significance of implicit cues and cross-modal anchoring, such as interpreting packaging measures in text and validating them with a visual scale. The efficiency profile indicates that one should select debate based on \(\Delta\mathrm{Acc}/\Delta t\) within a specified latency budget, rather than solely on absolute gain, as minor absolute improvements in robust models can be critical, while more substantial and time-efficient gains are characteristic of the weak models.

 We recommend a selective implementation approach with precise budget management for deployment.  Use a single-round debate by default and initiate a second-round debate only if necessary.  Implement an adaptive stopping criterion that ceases when the last iteration enhances accuracy by less than a user-defined threshold in relation to the latency budget.  In summary, Table~\ref{tab:debate-lat-acc} endorses a pragmatic operating point: a one-round debate achieves an optimal equilibrium between quality and cost for weak models, while simultaneously addressing complex cases for more robust models. Additionally, meticulous triggering and early cessation retain the majority of the advantages with minimal time investment.

\vspace{-0.5em}

\section{Conclusion}

In this study, we introduced \textsc{\modelname}, a multi-agent debate framework that enhances implicit AVE performance in multimodal e-commerce by integrating textual and visual reasoning across multiple MLLMs. Through a series of debate rounds, we discovered that agents collectively refine their responses in just one or two rounds of discussion, significantly improving inference compared to single MLLM approaches and previous baselines. Specifically, we observed that low-accuracy attributes benefit the most; however, excessive rounds can lead to confusion, underscoring the need to balance the number of debate iterations. Additionally, our experiments show that weaker agents substantially improve through interaction with stronger agents, often reaching performance comparable to that of the stronger agents. These findings highlight the potential of multi-agent debate strategies to overcome single MLLM limitations and provide scalable solutions for implicit AVE and related multimodal tasks. In the future, we plan to extend this approach to other e-commerce domains.

\newpage
\section{Limitations}

Our multi-agent debate framework, \textsc{\modelname}, demonstrates a strong capability for improving performance on the implicit AVE task compared to previous baseline methods and single MLLM approaches. However, our work has limitations. We compared our approach with only a limited number of baselines, as implicit AVE is a relatively new task and current research is still in its early stages. Additionally, many related works are not open-sourced or are unavailable. Therefore, we hope that \textsc{\modelname} can serve as a milestone for multimodal implicit AVE, providing a clear direction for future improvements.




\bibliography{custom}

\clearpage 

\appendix

\clearpage 
\vspace*{-1.5\baselineskip}
\begingroup
\makeatletter
\setcounter{tocdepth}{2}

\vspace*{-0.25em}
\begin{center}
  {\large\bfseries Appendix Contents}\par
  \vspace{0.35em}
  \hrule
  \vspace{0.55em}
\end{center}

\small
\renewcommand{\baselinestretch}{1.5}\selectfont
\setlength{\parskip}{0pt}
\setlength{\parindent}{0pt}
\def\@dotsep{1.2} 

\renewcommand*\l@section[2]{%
  \vspace{0.15em}%
  \@dottedtocline{1}{0em}{2.0em}{\bfseries #1}{\bfseries #2}%
}

\renewcommand*\l@subsection[2]{%
  \@dottedtocline{2}{1.25em}{2.6em}{\footnotesize #1}{\footnotesize #2}%
}

\@starttoc{apx}
\makeatother
\endgroup

\vspace{1.2em}
\hrule
\vspace{0.4em}

\counterwithin{figure}{section}
\counterwithin{table}{section}

\section{Converge Examples}
\label{sec:appendix1}

\begin{figure*}[t]
  \centering
  \includegraphics[width=\textwidth]{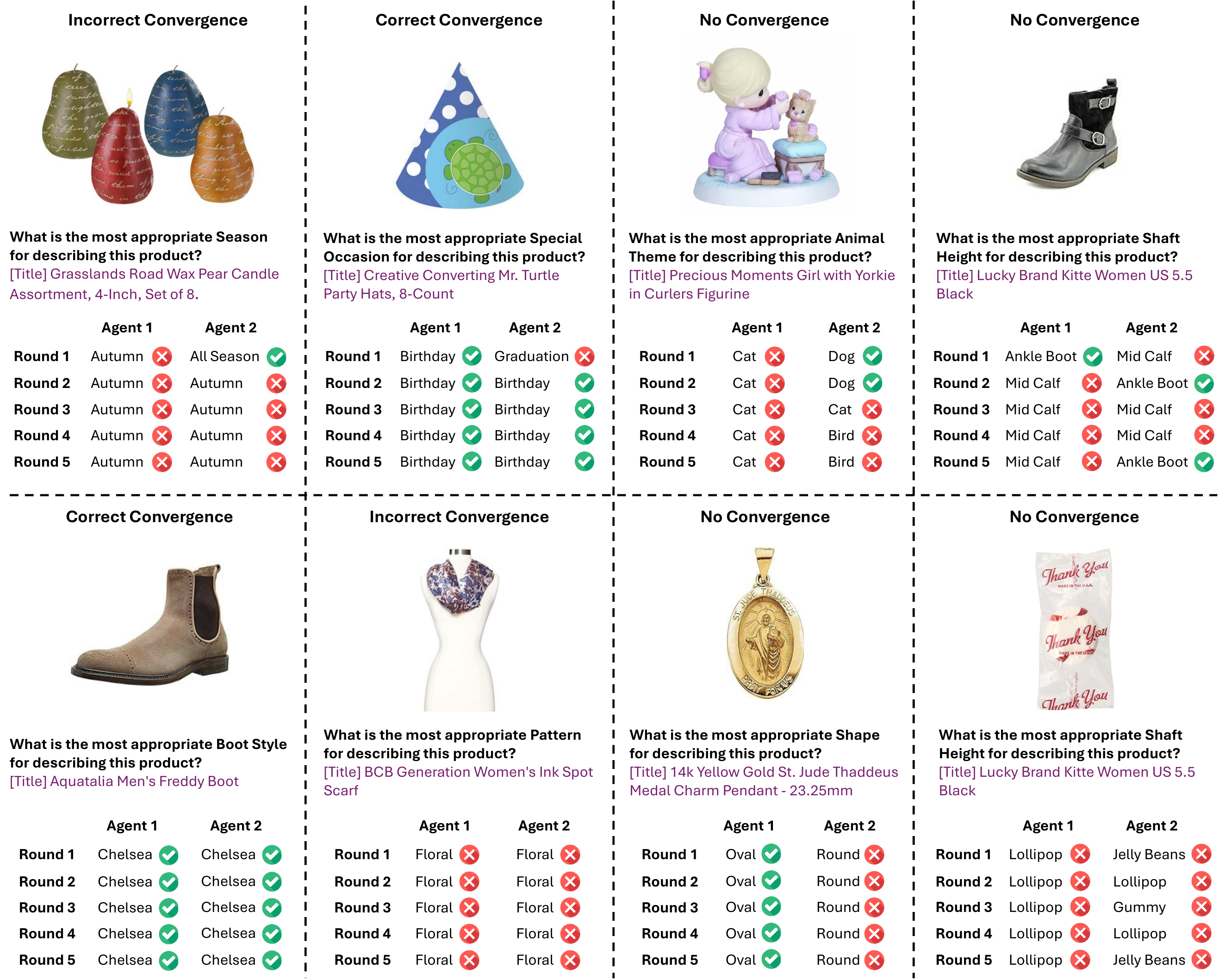}
  \caption{Illustration of correct, incorrect, and no convergence outcomes observed during agent debate.}
  \label{fig:fig2}
\end{figure*}

Our empirical findings indicate that multi-agent MLLM debates frequently reach consensus after just one or two rounds. We identified three distinct convergence outcomes, including correct convergence, incorrect convergence, and no convergence, as illustrated in Figure~\ref{fig:fig2}. Correct convergence occurs when both agents consistently maintain correct predictions or when they correct an initially incorrect prediction through debate. Incorrect convergence arises when the agents persist in erroneous predictions or transition from a correct to an incorrect inference. No convergence occurs if one or both agents repeatedly change their answers without reaching a stable agreement. Although our goal is to achieve correct convergence, the process by which consensus is reached remains uncertain. In fact, convergence behavior varies depending on the debate configuration (e.g., whether the debate involves identical agents, different agents, or varied input conditions).

Our observations further reveal that agents tend to incorporate and accept feedback from their counterparts across different scenarios. We believe this behavior is influenced by the underlying instruction tuning and reinforcement learning mechanisms governing the agents' reasoning processes \citep{ouyang2022training}. In many cases, agents prioritize the prompt input over their own initial stance.

\section{Model Selections}
Our experiments were completed before the public release of newer Qwen3-family models (e.g., Qwen3-VL and Qwen3-Omni). In addition, Qwen2.5-Omni and Qwen3-Omni are primarily optimized for real-time multimodal streaming (text, images, audio, video) and are generally comparable to same-sized Qwen2.5 or Qwen3 text or VL counterparts on standard benchmarks. As a sanity check, we additionally evaluated Qwen3-VL-7B on a subset of our tasks. Although the absolute performance improves slightly, the relative behavior across models and the qualitative trends discussed in the paper remain consistent. Therefore, replacing our backbone (\Qwenname) with Qwen2.5-Omni or Qwen3-VL or Omni would not materially change our main findings.

\section{Analysis and Potential Improvement}
\label{sec:appendix2}
According to the \textsc{\modelname} study, the multi-agent debate framework generally leads the models to agree on an answer within one or two rounds. In many of those cases, the agreed answer is indeed correct. Empirically, this is reflected in the rapid rise in F1 scores for most domains, as both agents adopt each other’s reasoning to correct initial mistakes. However, the study also documents instances of incorrect convergence, where both agents agree on the wrong attribute, because one agent’s flawed reasoning can unduly influence its partner. By the third round, the number of cases that worsened begins to exceed the number that improved, indicating that further rounds can sometimes push the consensus in the wrong direction. Finally, a small fraction of debates never reach a stable answer (no convergence), with agents repeatedly changing their responses without reaching an agreement.

To improve both the accuracy and stability of convergence, we suggest several refinements. First, introduce a stricter validation step during debates so that an agent only changes its answer when it can explicitly verify its peer’s justification, rather than simply deferring to it. Second, mitigate the teacher-student imbalance observed when a stronger model adopts a weaker model’s error by weighting updates according to model reliability, allowing the stronger model to critique without being forced to revise unless there is clear, corroborated evidence. Third, apply an early stopping criterion once consensus is reached or the F1 score gains plateau, since most benefits occur in the first two rounds. Extra rounds can introduce confusion. Finally, further instruction that tunes the agents toward a more skeptical debate style, encouraging them to maintain their initial answers unless the opposing argument is highly convincing. Together, these measures would help ensure that when agents agree, they are far more likely to converge on the correct inference and less likely to oscillate or jointly adopt wrong answers.

\subsection{Compatibility with Stronger Settings}
Importantly, \textsc{\modelname} is \emph{compatible} with stronger settings. One can directly plug in few-shot prompts, instruction-tuned backbones, or lightly fine-tuned LoRA variants as debate agents, and we expect the \emph{absolute} accuracy to increase. Meanwhile, our main qualitative findings about multi-agent debate. For example, that most gains occur within the first 1--2 rounds and that excessive debating can yield diminishing returns or drift, are expected to remain stable, since these effects primarily arise from interaction structure and information exchange rather than from a specific parameterization of the backbone. A full, controlled comparison across zero-shot, few-shot, and instruction-tuned variants would require additional large-scale training and careful curation of domain-labeled data (including coverage of long-tail profiles and continuously updated catalogs), which is beyond the scope of this paper. 

\subsection{Weighted Aggregation of Debate Agents}
\label{app:weighted_aggregation}

In the main paper, we aggregate agents' outputs with \emph{equal weighting}. This is a conscious design choice: it keeps \textsc{\modelname} simple, transparent, and broadly applicable across domains, without introducing extra assumptions about which agent should be trusted a priori. Nevertheless, \emph{weighted} aggregation could potentially further improve performance by exploiting heterogeneity among agents and arguments.

A natural extension is to assign higher influence to ``expert'' agents, such as (i) stronger backbones, (ii) agents specialized in different evidence modalities (e.g., visual vs.\ textual grounding), or (iii) agents that consistently provide higher-quality arguments on specific task subsets. Another complementary direction is \emph{confidence- or evidence-aware} weighting, where arguments supported by stronger evidence (e.g., explicit citations to relevant product attributes, clearer visual cues, or internally consistent reasoning) or higher calibrated confidence receive more weight in the final decision.

Systematically studying weighting strategies requires an additional learning and calibration layer---including how to estimate confidence, how to validate ``expertise'' across tasks, and how to prevent over-reliance on spurious signals. This introduces new design choices and hyperparameters, and it requires careful validation to ensure robustness under distribution shifts (e.g., new product categories or evolving catalogs). Because our goal is to evaluate \textsc{\modelname} as a general debate framework with minimal supervision, we leave this exploration beyond the scope of the current paper.

We view at least three directions as particularly promising future work: (i) \textbf{expertise-aware weighting}, where per-agent weights are determined by backbone strength or specialization; (ii) \textbf{confidence- or evidence-aware weighting}, where weights depend on calibrated uncertainty or measurable evidence quality; and (iii) \textbf{learning debate weights}, where the weighting policy (or even the debate graph or structure) is treated as a reinforcement learning problem that optimizes downstream task performance subject to constraints such as cost or debate length. We will clarify these points in the revised version and explicitly state that equal weighting is a deliberate baseline for generality, while weighted aggregation offers a principled avenue for further improvements.

\setlength{\tabcolsep}{4pt} 


\section{Debate Scripts }
\label{app:debate-scripts}

\noindent In this appendix, we present concise multimodal debate traces and scripts between \texttt{\llamaname} (Agent~1) and \texttt{\phiname} (Agent~2). Each case begins with a brief summary, then shows an \emph{Inference Round} followed by a \emph{Debate Round}. The scripts illustrate how agents integrate textual cues (e.g., “Room Darkening,” measurements) with visual evidence (scene context, shape/fit).

\setlength{\fboxsep}{6pt}
\setlength{\fboxrule}{0.4pt}

\newcommand{\DebateAgent}[4]{%
  \textbf{#1} (\emph{#2})\\
  \textsc{Answer:} \emph{#3}\\
  \textsc{Explanation:} #4\par
}

\newcommand{\RoundBox}[2]{%
  \noindent
  \fbox{%
    \parbox{\dimexpr\linewidth-2\fboxsep-2\fboxrule\relax}{%
      \textbf{#1}\par\medskip
      \raggedright #2%
    }%
  }%
}

\newcommand{\CaseHeader}[1]{%
  \vspace{0.75em}
  \noindent\textbf{#1}\par
  \vspace{0.25em}\hrule\vspace{0.5em}
}

\subsection{Case A — Curtain Placement}
\noindent In Case~A, the agents weigh product text (the ``Room Darkening'' cue, which typically signals bedroom use for blocking light) against scene context (a sofa and coffee table implying a living room). The purpose is to frame the cross-modal tension so the reader knows the lens for evaluation; after integrating both modalities, the debate converges on \textbf{Bedroom}.

\RoundBox{Inference Round}{%
  \DebateAgent{\llamaname}{Agent 1}{Bedroom}{The title explicitly mentions ``Room Darkening,'' typically for sleeping spaces. The image shows curtains near a headboard and nightstand, which are common bedroom elements.}
  \medskip
  \DebateAgent{\phiname}{Agent 2}{Living Room}{A sofa and coffee table appear in the background, suggesting a living-room placement.}
}

\vspace{0.5em}
\RoundBox{Debate Round}{%
  \DebateAgent{\llamaname}{Agent 1}{Bedroom}{``Room Darkening'' is uncommon in living rooms where natural light is preferred. Listing metadata emphasizes ``master bedroom'' usage.}
  \medskip
  \DebateAgent{\phiname}{Agent 2}{Bedroom}{Re-checking text and image supports bedroom as the intended location.}
}

\subsection{Case B — Pendant Shape}
\noindent In Case B, visual inspection suggests a circle, but the summary signals that careful multimodal grounding, combining the title’s \emph{23.25\,mm} numeric detail with visible height-to-width proportion, drives the outcome. This primes the reader to look for numeric–visual alignment during the debate, which ultimately converges on \textbf{Oval}.

\RoundBox{Inference Round}{%
  \DebateAgent{\llamaname}{Agent 1}{Oval}{Height exceeds width, yielding an elongated silhouette. The 23.25\,mm figure in the title denotes the longer axis.}
  \medskip
  \DebateAgent{\phiname}{Agent 2}{Round}{At a glance, the pendant appears nearly circular without precise measurement.}
}

\vspace{0.5em}
\RoundBox{Debate Round}{%
  \DebateAgent{\llamaname}{Agent 1}{Oval}{Compared with the model’s finger, the medal is taller than wide. A single numeric value indicates length, not diameter.}
  \medskip
  \DebateAgent{\phiname}{Agent 2}{Oval}{Combining the visual ratio with the title’s measurement confirms an oval shape.}
}

\subsection{Case C — Footwear Category: Overtrust Example}

\noindent This section highlights the \emph{failure mode} the case is designed to illustrate: a stronger agent can be impacted by a confident but modality-biased claim. The initial evidence supports \textbf{Ankle Boot} (shaft at ankle level), yet during debate the stronger model flips to \emph{Knee High} after over-weighting a visual cue. The purpose is to warn that, in multimodal debate, assertive but unbalanced reasoning can degrade accuracy—here, the correct label remains \textbf{Ankle Boot}. \\

\RoundBox{Inference Round}{%
  \DebateAgent{\llamaname}{Agent 1}{Ankle Boot}{The shaft stops just above the ankle bone, a hallmark of ankle boots.}
  \medskip
  \DebateAgent{\phiname}{Agent 2}{Knee High}{Styling suggests coverage up to mid-calf, interpreted as knee-high.}
}

\vspace{0.5em}
\RoundBox{Debate Round}{%
  \DebateAgent{\llamaname}{Agent 1}{Knee High}{On re-check, the shaft appears to extend well above the ankle, suggesting knee-high.}
  \medskip
  \DebateAgent{\phiname}{Agent 2}{Ankle Boot}{Revisiting cues shows the shaft sits at ankle level; ``Ankle Boot'' is correct.}
}
\vspace{3mm}

\noindent These case-specific summaries orient readers to the evidence types and intended lesson \emph{before} the transcripts: (A) text vs.\ scene context with convergence on bedroom use; (B) numeric measurement plus visual proportion resolving shape ambiguity; (C) overtrust of a single modality leading to an incorrect flip. Used this way, summaries make the ensuing debate easier to scan and evaluate.



\section{Debate Stability}
\label{app:ave-stability}

Table~\ref{tab:std-table} reports overall accuracy as mean ± SD for a single round and a one round debate. A capacity effect is evident: weaker models show larger variability and larger average gains, while stronger models remain comparatively stable and still improve. To make variability scale invariant, we examine the coefficient of variation (SD divided by mean). For \texttt{\llamaname}, the relative SD drops from about 0.70\% to 0.60 \%; for \texttt{GPT-4o}, it drops from about 0.55\% to 0.45\%; for \texttt{GPT-o1}, it drops from about 0.58\% to 0.50\%. In contrast, \texttt{\phiname} rises from about 5.70\% to 7.31\%, \texttt{\Qwenname} rises from about 1.09\% to 1.74\%, and \texttt{Claude-3.5-Sonnet} rises from about 0.56\% to 1.12\%. Thus, debate tends to reduce relative variability for the stronger trio and increase it for the weaker model. There is also an association between mean lift and change in SD: the largest gains, such as \texttt{\phiname} at +7.24\% and \texttt{\Qwenname} at +5.16 \%, coincide with higher SDs, which suggests that debate introduces productive exploration for weaker systems at the cost of stability.

Debate consistently raises the performance for all models, and its effect on variability follows model capacity. For strong systems, debate both improves accuracy and slightly stabilizes outcomes, which is attractive for production settings that value predictable behavior. For weaker systems, debate yields larger average gains but can increase run to run spread, so it is best used with safeguards \citep{huang2025deepresearchguard}. In practice we recommend routing by expected efficiency and variance: enable a round of debate when the predicted gain per added second exceeds a latency budget and when the historical variance on the same slice is acceptable. Use fixed prompts and decoding settings, log seeds, and consider a temperature schedule that cools later rounds to limit variance growth. 

\begin{table*}[t]
\centering
\small
\setlength{\tabcolsep}{6pt}
\renewcommand{\arraystretch}{1.12}
\caption{Overall accuracy stability: mean ± SD (\%) for initial single round and one round debate. $\Delta$Mean is debate minus initial; $\Delta$SD is debate SD minus initial SD.}
\label{tab:std-table}
\resizebox{\textwidth}{!}{%
\begin{tabular}{l c c c c}
\toprule
\textbf{Model} &
\textbf{Initial Round (mean ± SD)} &
\textbf{Debate Round (mean ± SD)} &
\textbf{$\Delta$Mean (\%)} &
\textbf{$\Delta$SD} \\
\midrule
\llamaname & $76.17 \pm 0.53$ & $77.97 \pm 0.47$ & $+1.80$ & $-0.06$ \\
\phiname       & $46.68 \pm 2.66$ & $53.92 \pm 3.94$ & $+7.24$ & $+1.28$ \\
\Qwenname                   & $70.66 \pm 0.77$ & $75.82 \pm 1.32$ & $+5.16$ & $+0.55$ \\
GPT-4o                        & $85.21 \pm 0.47$ & $87.52 \pm 0.39$ & $+2.31$ & $-0.08$ \\
Claude-3.5-Sonnet             & $76.64 \pm 0.43$ & $78.58 \pm 0.88$ & $+1.94$ & $+0.45$ \\
GPT-o1                        & $84.47 \pm 0.49$ & $87.77 \pm 0.44$ & $+3.30$ & $-0.05$ \\
\bottomrule
\end{tabular}%
}
\end{table*}

\section{Attribute-Level Result}
\label{sec:attribute-level result}
\begin{figure*}[t]

  \includegraphics[width=\textwidth]{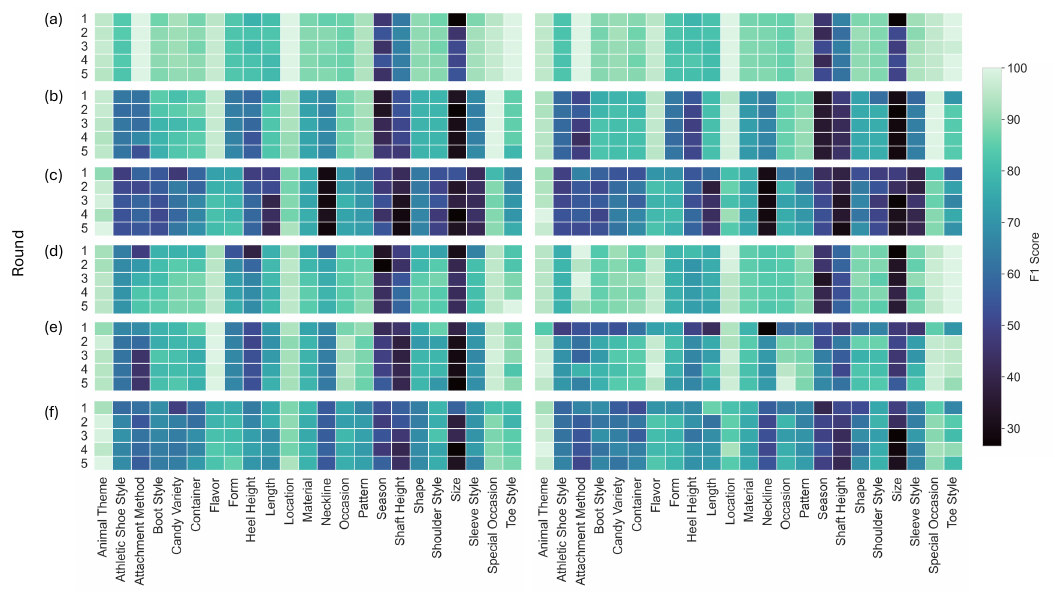}
  \caption{Attribute-level heatmap of the F1 score changes during the debate process for (a) Scenario 1: GPT-4o debating with GPT-4o; (b) Scenario 2: \llamaname~debating with \llamaname; (c) Scenario 3: \phiname~debating with \phiname; (d) Scenario 4: \llamaname~debating with GPT-4o; (e) Scenario 5: \llamaname~debating with \phiname; (f) Scenario 6: \Qwenname~debating with \Qwenname.}
  \label{fig:Attribute-Level Result}
\end{figure*}

We present the attribute-level results in Figures~\ref{fig:Attribute-Level Result}. In the same MLLM debate in Figures~\ref{fig:Attribute-Level Result}(a), \ref{fig:Attribute-Level Result}(b), \ref{fig:Attribute-Level Result}(c), and \ref{fig:Attribute-Level Result}(f), the attributes \emph{Size}, \emph{Season}, and \emph{Shaft Height} have great improvement across all attributes. In several scenarios, these underperforming attributes show marked improvements through the debate process. For instance, when \texttt{GPT-4o} debates with itself, the \emph{Size} attribute improves from 26.55\% to 54\% after four rounds of debate. However, during four rounds of debate, \texttt{\llamaname} sees the \emph{Size} attribute decrease from 40\% to 38.48\%, and \texttt{\phiname} experiences a decline in the F1-score for \emph{Size} from 37.8\% to 13.87\%. A similar pattern, though with more modest changes, is observed for the \emph{Season} and \emph{Shaft Height} attributes. These findings suggest that debate-induced improvements are concentrated on attributes with lower initial accuracy, which likely require diverse viewpoints and enhanced reasoning to achieve better inference. However, these improvements are only realized when the debating agent possesses sufficient knowledge. If an agent misinterprets the product image or text, its flawed reasoning may negatively influence its own performance and that of its debate partner in subsequent rounds, ultimately further degrading performance on low-accuracy attributes, as observed with \texttt{\phiname} and \texttt{\Qwenname}.

A similar pattern is also observed in the cross MLLM debates shown. In Figure~\ref{fig:Attribute-Level Result}(d), the left-hand side represents \texttt{\llamaname} while the right-hand side represents \texttt{GPT-4o}. In this scenario, \texttt{\llamaname} exhibits significant improvements in several attributes; for example, \emph{Attachment Method}, \emph{Form}, \emph{Heel Height}, and \emph{Neckline} improve by 30.39\%, 19.77\%, 25.6\%, and 15.32\%, respectively. These gains indicate that the weaker agent benefits considerably from the debate, particularly on attributes with lower initial accuracy. In contrast, \texttt{GPT-4o} experiences a slight decline in certain attributes: \emph{Shape}, \emph{Length}, and \emph{Shaft Height} decrease by 0.76\%, 1.17\%, and 3.32\%, respectively. Although \texttt{GPT-4o}'s negative changes are less pronounced than the positive improvements observed in \texttt{\llamaname}, the attributes that show the largest gains for the weaker agent do not necessarily correspond to those with the most significant declines for the stronger agent, indicating that these effects are not directly correlated.

Similarly, in the same debate settings between \texttt{\llamaname} and \texttt{\phiname}, as shown in Figure~\ref{fig:Attribute-Level Result}(e). In this configuration, \texttt{\phiname}'s inference improves markedly on several attributes—\emph{Neckline}, \emph{Shape}, \emph{Candy Variety}, and \emph{Sleeve Style} increase by 53.01\%, 35.99\%, 35.92\%, and 42.78\%, respectively. In contrast, some attributes for \texttt{\llamaname} decline, with \emph{Size}, \emph{Attachment Method}, and \emph{Shoulder Style} decreasing by 11.12\%, 20.49\%, and 13.52\%. These findings suggest that low-accuracy attributes are particularly sensitive to the quality of reasoning: when accurate reasoning is applied, these attributes benefit significantly, but flawed reasoning, even from a stronger agent, can lead to notable performance decrements.

\section{Use of LLMs}

We used LLMs only to refine wording (grammar, clarity, and style) of text. It was not used to generate ideas, write substantive content, design experiments, analyze data, or produce results. No non-public data were provided to the model, and all content was verified and finalized by the authors.

\end{document}